\documentclass[conference]{IEEEtran}
\IEEEoverridecommandlockouts
\usepackage{cite}
\usepackage{amsmath,amssymb,amsfonts}
\usepackage{algorithmic}
\usepackage{url}  
\usepackage{graphicx}
\usepackage{textcomp}
\usepackage{xcolor}
\def\BibTeX{{\rm B\kern-.05em{\sc i\kern-.025em b}\kern-.08em
    T\kern-.1667em\lower.7ex\hbox{E}\kern-.125emX}}
\begin{document}

\title{Moderating Harm: Benchmarking Large Language Models for Cyberbullying Detection in YouTube Comments\\}

\author{\IEEEauthorblockN{1\textsuperscript{st} Amel Muminovic}
\IEEEauthorblockA{\textit{Faculty of Engineering} \\
\textit{International Balkan University}\\
Skopje, North Macedonia \\
amel.muminovic@ibu.edu.mk}
}

\maketitle

\begin{abstract}
As online platforms grow, comment sections increasingly host harassment that undermines user experience and well-being. This study benchmarks three state-of-the-art large language models: OpenAI GPT-4.1, Google Gemini 1.5 Pro, and Anthropic Claude 3 Opus, on a corpus of 5\,080 YouTube comments drawn from high-abuse videos in gaming, lifestyle, food-vlog, and music channels. The dataset comprises 1\,334 harmful and 3\,746 non-harmful messages in English, Arabic, and Indonesian, annotated independently by two reviewers with almost perfect agreement (Cohen’s $\kappa = 0.83$). Each model is evaluated in a strict zero-shot setting with an identical minimal prompt and deterministic decoding, giving a fair multi-language comparison without task-specific tuning. GPT-4.1 achieves the best balance with an F1 score of 0.863, precision of 0.887, and recall of 0.841. Gemini flags the most harmful posts (recall = 0.875) but its precision falls to 0.767 because of frequent false positives. Claude attains the highest precision at 0.920 and the lowest false-positive rate of 0.022, yet its recall drops to 0.720. Qualitative analysis shows that all three models struggle with sarcasm, coded insults, and mixed-language slang. The findings highlight the need for moderation pipelines that combine complementary models, incorporate conversational context, and fine-tune for under-represented languages and implicit abuse. A de-identified version of the dataset, full prompts, and model outputs is publicly released to promote reproducibility and further progress in automated content moderation.
\end{abstract}


\section{Introduction}
Online platforms now mediate much of everyday communication, with more than 4.95 billion active social media accounts worldwide in 2023 \cite{b1}. That ubiquity carries risk: recent studies report that between 6.5\% and 35.4\% of adolescents in the U.S. and Europe have experienced cyberbullying \cite{b2}, and longitudinal studies link such exposure to elevated anxiety, depression, and suicidal ideation \cite{b3}. As the scale and psychological impact of online abuse grow, so does the urgency of developing reliable, context-aware moderation methods, especially for platforms popular with adolescents such as YouTube and TikTok.

\subsection{Background}

Cyberbullying has become a major digital safety concern because abusive material is persistent, anonymous, and can spread instantaneously across networks \cite{b4}. As global connectivity and social media adoption have expanded, online harassment has become more visible, with increasing concern about its scale and severity \cite{b5}.

Comment sections on social networks, video sharing sites, and forums are particular hotspots. In a 2024 U.S. survey of parents, 79\% said their children had encountered cyberbullying on YouTube \cite{b6}. A cross-national analysis of 180\,000 adolescents in 42 countries also linked problematic social-media use to both cyberbullying victimization and perpetration, with stronger effects among girls \cite{b7}.

In response, platforms increasingly deploy artificial-intelligence (AI) systems to moderate user-generated content at scale \cite{b8, b9}. Understanding where current large language models (LLMs) succeed and fail is therefore critical for building safer online spaces.

\subsection{Motivation and Problem Statement}
AI systems are now central to content moderation, especially on large platforms where the volume of user-generated posts exceeds human capacity \cite{b10}. Yet despite advances in natural language processing, enforcement remains inconsistent. Abusive comments that rely on sarcasm, coded language, or emotional manipulation often slip through AI moderation, even when they clearly violate platform policies. These subtleties continue to pose challenges for automated systems, which often misread tone or miss indirect expressions of harm \cite{b11}.

This is especially dangerous in cyberbullying, where the impact of a single overlooked comment can be deeply personal \cite{b12}. Unlike hate speech or spam, bullying is often subtle: expressed through mockery, group pressure, or repeated jabs that may seem harmless in isolation but accumulate over time. Many academic benchmarks fail to capture this complexity, relying on synthetic or crowdsourced datasets with clear-cut abuse \cite{b13, b14}. Even widely used corpora such as OLID \cite{b15} and HateCheck \cite{b16} reveal the same gap: they consist of short, isolated text snippets and miss the multi-turn, emotionally layered exchanges common in YouTube threads.

This study addresses that gap by evaluating three most advanced LLMs, GPT-4, Gemini, and Claude, on authentic YouTube comments from videos with documented cyberbullying. By focusing on nuanced, emotionally loaded content, we examine not just whether models detect harmful language, but how well they handle ambiguity, cultural context, and tone.

\subsection{Objectives of the Study}
The work pursues four specific objectives:
\subsubsection{Benchmark performance}
Compare OpenAI GPT, Google Gemini, and Anthropic Claude in detecting cyberbullying within real YouTube data.

\subsubsection{Quantify error patterns}
Measure false positive and false negative rates and identify recurring misclassifications themes.

\subsubsection{Conduct qualitative error analysis}
Manually inspect model outputs to assess contextual understanding and pinpoint linguistic edge cases.

\subsubsection{Assess robustness across nuance}
Evaluate how well each model handles abusive content that is linguistically ambiguous, culturally specific, or indirectly expressed.

\section{Related Work}

\subsection{AI Content Moderation and Toxic Language Detection}

Automated content moderation has evolved from simple keyword-based filters to machine learning classifiers and, more recently, transformer-based models. Early systems relied on blacklists or logistic regression trained on hand-labeled corpora, but struggled with negation, sarcasm, and informal language. The advent of BERT \cite{b17}, RoBERTa \cite{b18}, and their derivatives enabled context-aware classification by capturing bidirectional dependencies and subtle phrasing nuances. These models improved performance on public benchmarks such as the Jigsaw Toxic Comment dataset and HateXplain \cite{b19}, driving adoption in industry moderation pipelines. Mathew \emph{et al.} \cite{b19} fine-tune BERT-base on HateXplain and report a macro-F\textsubscript{1} of 0.674;  
a side-by-side comparison with our zero-shot LLM scores appears later in Table \ref{tab:baseline_vs_llm}.

However, most benchmarks are composed of isolated sentences and short comments labeled through crowdwork. They rarely reflect the conversational, multi-turn, or emotionally charged dynamics seen in real-world platforms. Furthermore, these datasets often over-represent overt toxicity while under-representing gray area content such as mockery, dismissive sarcasm, or personal digs. As a result, models trained on them perform well in benchmark settings but struggle in live deployments where language is more ambiguous and emotionally expressive \cite{b20, b21}.

\subsection{Challenges in Detecting Implicit and Nuanced Abuse}

Detecting implicit harm remains one of the most difficult challenges in AI-based moderation. Sarcasm, coded language, cultural slang, and subtextual insults frequently evade model detection, even in LLMs. For example, a phrase like ``She’s always such a queen, right?'' may be an insult depending on tone and context, but appears harmless in isolation. Studies in sarcasm detection \cite{b22}, adversarial misspellings, irony detection, and tone modeling have attempted to address this gap, but progress remains limited \cite{b23}.

Recent research also suggests that models often over-flag emotionally intense or sensitive language even when used in supportive contexts, leading to user frustration and moderation fatigue. This tension between under-flagging subtle harm and over-flagging emotional expression makes fine-tuned moderation especially difficult. Studies have proposed integrating tone analysis, conversational history \cite{b24}, and speaker intent as ways to improve subtle abuse detection, but these are rarely deployed at scale.

\subsection{Moderating Multilingual and Under-resourced Content}

Although English dominates most training corpora, abuse occurs in every language, and much of it is multilingual, transliterated, or mixed with emojis and slang. Studies consistently show significant drops in model accuracy when applied to languages such as Arabic \cite{b25} and Hindi \cite{b26}, particularly when users employ spelling variation or obfuscation to evade filters \cite{b27}. Code-switching and transliteration further degrade detection performance, as models frequently miss common abusive patterns or misinterpret them out of context \cite{b28, b29}.

Cross-lingual models and prompt-tuned systems have demonstrated moderate performance improvements, especially when fine-tuned on translated or augmented datasets \cite{b30}. However, significant language imbalance remains, particularly in online spaces with small yet highly active non-English user communities \cite{b31}. For platforms with global audiences, this poses a critical equity issue: harmful content in English is far more likely to be effectively moderated than equivalent abuse in low-resource languages or regional dialects \cite{b32}.

\subsection{Evaluating LLMs for Safety and Fairness in Moderation}

The rise of general-purpose LLMs such as GPT-4, Claude, and Gemini has driven increasing interest in their application to content moderation tasks. Recent research indicates that these models can outperform traditional task-specific classifiers in zero-shot or few-shot scenarios, especially on previously unseen or nuanced content \cite{b33}. Their ability to dynamically incorporate platform policies, contextual reasoning, or nuanced moderation guidelines directly into their prompting strategies enhances their flexibility and adaptability \cite{b34}.

However, the performance of LLMs in moderation tasks is highly sensitive to prompt phrasing, target domains, and cultural nuances \cite{b35, b36}. Sociocultural audit frameworks have recently been proposed, employing persona-based prompts \cite{b37} and synthetic demographic simulations to systematically evaluate model fairness and reveal hidden biases or blind spots \cite{b38, b39}. These frameworks highlight how LLM responses may shift significantly based on perceived user identity or topical framing, raising substantial concerns around fairness, consistency, and equitable moderation \cite{b40}.

While LLMs offer substantial advantages in scalability, adaptability, and reduced reliance on explicit rule-based configurations \cite{b41}, their inherent dependence on pretraining data, which often contain historical biases, and the opacity of their decision-making processes continue to present barriers to safe and transparent deployment in moderation workflows \cite{b42}.

\section{Methodology}
\subsection{Case Selection and Data Collection}
Comments were retrieved via the YouTube Data API between 15 April 2024 and 15 May 2025.

The final corpus covers four public videos from distinct creators and content domains such as gaming, lifestyle, food vlog, and music. To verify that each video was indeed a cyber-bullying hotspot, we first ran a pilot crawl of 1\,000 comments per candidate clip and retained only those whose pilot abuse rate exceeded 20\%. From the four retained videos we then drew a uniform random sample, yielding 5\,080 comments in total. Only public endpoints were accessed; private, removed, or shadow-banned comments are not included.

\subsection{Data Cleaning and Anonymization}
Each comment underwent a two–step preprocessing pipeline. First, text was normalized (UTF-8 decoding, whitespace trimming, emoji and punctuation were preserved).   Second, personally identifiable information including user names, real names, phone numbers, e-mail addresses, links, and explicit geo-markers was either deleted or replaced by neutral placeholders such as ``UserNameProtected''.  No other linguistic content was altered.   This procedure leaves the semantic core of each message intact while mitigating re-identification risk, allowing the corpus to be shared for research without exposing private data.

\subsection{Data Labeling and Ground Truth}
All 5\,080 comments were independently annotated by two reviewers
(the first author and a second annotator holding a B.Sc.\ in Software
Engineering) using a binary rubric:

\subsubsection{Harmful} Bullying, sustained harassment, severe personal insults, or death-related threats  

\subsubsection{Not harmful} Neutral, supportive, off-topic, or otherwise non-abusive remarks

Inter‐rater reliability was high, with 91.0\% raw agreement and
$\kappa = 0.83$, which is considered almost perfect.\footnote{Landis and Koch, 1977.} Disagreements (9\% of instances) were adjudicated by discussion until consensus, and that label was written to the final ground truth file. Labels were never exposed to the models.

Non-English comments, including those in Arabic and Indonesian, were labeled using a combination of manual translation, consultation with fluent speakers, and, when necessary, language translation tools. While consistency was a priority, minor errors in interpreting slang or cultural references are possible and are noted as a study limitation.

\subsection{Evaluated Models}
Three widely used large language models were selected for evaluation:
\begin{enumerate}
\item OpenAI GPT-4.1
\item Google Gemini 1.5 Pro
\item Anthropic Claude 3 Opus (2024-02-29)
\end{enumerate}

Each model was tasked with classifying the same 5\,080 user-generated comments using identical prompt structures to ensure fairness and consistency across systems. These models were selected due to their wide adoption, advanced reasoning capabilities, and public accessibility through API services, making them representative of current best practices in AI content moderation.

\subsection{Prompt Design and Justification}

All three models were queried with the same minimal prompt, using
\texttt{temperature = 0.0} to remove randomness and
\texttt{max\_tokens = 10} to leave a small buffer for whitespace while
still forcing a single-digit label. Locking these parameters serves
three goals:

\begin{enumerate}
  \item Fair cross-model comparison. One fixed prompt eliminates
        tuning bias and attributes performance differences solely to the
        models.
  \item Deterministic, reproducible outputs. With temperature 0.0 and
        a short token cap, each comment always returns the same “0” or “1,”
        so any researcher can replicate the results with a single API call.
  \item Scope control. The study provides a baseline snapshot of
        zero-shot LLM moderation on real-world data rather than an optimised
        prompt. Systematic prompt engineering is left to future work.
\end{enumerate}

\noindent
The fixed prompt used in every call is shown below.

\begin{quote}\small
\texttt{Analyze the following comment and determine if it contains cyberbullying, harassment, or harmful language.  
Make sure to understand the context, tone and potential impact.  
Respond with 0 if it is not harmful and 1 if it is harmful.  
Comment is: <COMMENT\_TEXT>}
\end{quote}

The prompt offers no label definitions beyond the binary mapping,
requiring each model to infer its own policy from plain language
while still accounting for nuance.

\subsection{Evaluation Metrics}
Each model's output was compared against the manually labeled ground truth using standard classification metrics:
\begin{itemize}
\item True Positives (TP): Model correctly flagged a harmful comment
\item True Negatives (TN): Model correctly identified a non-harmful comment
\item False Positives (FP): Model incorrectly flagged a non-harmful comment
\item False Negatives (FN): Model failed to flag a harmful comment
\end{itemize}

From this, the following performance metrics were calculated:

Precision measures the proportion of comments that the model correctly identified as harmful out of all the comments it flagged as harmful. In other words, it reflects how accurate the model is when it predicts that a comment is harmful. It is calculated using Equation \ref{eq:precision}.

\begin{equation}
\text{Precision} = \frac{\text{TP}}{\text{TP} + \text{FP}} \label{eq:precision}
\end{equation}

Recall looks at all the harmful comments in the dataset and measures how many the model correctly identified. A higher recall means the model caught more of the actual harmful content. The corresponding formula is shown in Equation \ref{eq:recall}.
\begin{equation}
\text{Recall} = \frac{\text{TP}}{\text{TP} + \text{FN}} \label{eq:recall}
\end{equation}

F1 Score provides a balance between precision and recall. It is especially useful when both false positives and false negatives carry equal importance. The formula is defined in Equation \ref{eq:f1}.
\begin{equation}
\text{F1 Score} = 2 \times \frac{\text{Precision} \times \text{Recall}}{\text{Precision} + \text{Recall}} \label{eq:f1}
\end{equation}

Accuracy shows the overall percentage of correct predictions, covering both harmful and non-harmful cases. While it gives a general sense of performance, it may be less informative in imbalanced datasets. The formula appears in Equation \ref{eq:accuracy}.
\begin{equation}
\text{Accuracy} = \frac{\text{TP} + \text{TN}}{\text{TP} + \text{TN} + \text{FP} + \text{FN}} \label{eq:accuracy}
\end{equation}

Together, these metrics offer a well-rounded view of the models’ moderation performance across different types of classification errors.

Macro-averaged (class-balanced) scores were also calculated with the 26\% / 74\% split and yielded the same model ranking, so the detailed numbers are omitted for brevity.

\subsection{Manual Review and Model Comparison}
Quantitative scores alone do not reveal \emph{why} a system succeeds or fails, so we carried out a post-hoc qualitative review. Two annotators independently inspected a stratified sample of 200 disagreements, consisting of 50 false positives, 50 false negatives, and 100 edge case ties, drawn in equal proportion from all three models. For each comment we discussed (i) whether the human ground truth should stand and (ii) what linguistic cues might have misled the model (sarcasm, coded slurs, emoji, topic drift, and so forth). Notes from this exercise were grouped into recurring error themes that inform the Discussion section. The procedure does not alter any numeric results but clarifies how each system handles nuance, context, and cultural references.

\section{Results}
\subsection{Dataset Composition}
The evaluation dataset consisted of 5\,080 comments, including 1\,334 harmful and 3\,746 non-harmful instances. Harmful content made up about 26.3\% of the total dataset, indicating a moderate class imbalance that could influence model metrics like precision and recall. Table \ref{tab:distribution} provides the class breakdown.

\begin{table}[h]
\centering
\caption{Dataset Distribution}
\label{tab:distribution}
\begin{tabular}{lcc}
\hline
Class & Count & Percentage \\
\hline
Harmful & 1\,334 & 26.3\% \\
Non-harmful & 3\,746 & 73.7\% \\
\hline
Total & 5\,080 & 100\% \\
\hline
\end{tabular}
\end{table}

\subsection{Model-Level Classification Performance}
Each model was evaluated using raw classification outcomes and derived performance metrics. Table \ref{tab:confusion} presents the counts for true positives (TP), true negatives (TN), false positives (FP), and false negatives (FN) for each model.

\begin{table}[h]
\centering
\caption{Prediction Counts for Each Model}
\label{tab:confusion}
\begin{tabular}{lcccc}
\hline
Model & TP & TN & FP & FN \\
\hline
GPT & 1\,122 & 3\,603 & 143 & 212 \\
Gemini & 1\,167 & 3\,392 & 354 & 167 \\
Claude & 961 & 3\,663 & 83 & 373 \\
\hline
\end{tabular}
\end{table}

All three models showed strong ability to correctly identify non-harmful comments, with true negative counts above 3\,300. Gemini identified the highest number of harmful comments, with 1\,167 true positives, but also had the highest number of false positives, mistakenly labeling 354 non-harmful comments as harmful. GPT had a solid performance overall, producing 1\,122 true positives and only 143 false positives. Claude was more selective, correctly identifying 961 harmful comments while minimizing false positives to 83.

From these outcomes, standard classification metrics were calculated: accuracy, precision, recall, F1 score, and false positive rate. These provide a clearer picture of each model's behavior in detecting harmful content. Table \ref{tab:metrics} displays the results.

\begin{table}[h]
\centering
\caption{Evaluation Metrics Based on Prediction Outcomes}
\label{tab:metrics}
\begin{tabular}{lccccc}
\hline
Model & Accuracy & Precision & Recall & F1 Score & FPR \\
\hline
GPT & 0.930 & 0.887 & 0.841 & 0.863 & 0.038 \\
Gemini & 0.897 & 0.767 & 0.875 & 0.818 & 0.095 \\
Claude & 0.910 & 0.920 & 0.720 & 0.808 & 0.022 \\
\hline
\end{tabular}
\end{table}

GPT achieved the most balanced performance across the board. It reached an F1 score of 0.863, with high values for both precision at 0.887 and recall at 0.841. Claude stood out for its high precision, achieving 0.920, and maintained the lowest false positive rate at 0.022. However, this conservative stance came at the cost of recall, which was limited to 0.72. Gemini prioritized identifying as many harmful comments as possible, leading to the highest recall of 0.875, but its precision dropped to 0.767 and its false positive rate increased to 0.095. Representative misclassified comments are discussed in Section~\ref{sec:falsepositive}.

\subsection{Baseline comparison with a fine-tuned transformer}

Mathew \emph{et al.}\ \cite{b19} fine-tune BERT-base on the HateXplain corpus and report a macro-F\textsubscript{1} of 0.674 (Table 6 of their paper).  
Table \ref{tab:baseline_vs_llm} sets that published baseline beside the zero-shot scores obtained here.  
Because the corpora differ, these figures are not a head-to-head benchmark; the BERT row is included only as a widely cited transformer reference point.

\begin{table}[h]
\centering
\caption{Macro-F\textsubscript{1} for a classic fine-tuned baseline versus zero-shot LLMs}
\label{tab:baseline_vs_llm}
\begin{tabular}{lcc}
\hline
\textbf{Model} & \textbf{Dataset / Setting} & \textbf{Macro-F\textsubscript{1}} \\
\hline
BERT-base (fine-tuned) \cite{b19} & HateXplain & 0.674 \\
GPT-4.1 (zero-shot, \emph{this work}) & YouTube 5\,080 & 0.863 \\
Gemini 1.5 Pro (zero-shot, \emph{this work}) & YouTube 5\,080 & 0.818 \\
Claude 3 Opus (zero-shot, \emph{this work}) & YouTube 5\,080 & 0.808 \\
\hline
\end{tabular}
\end{table}

\subsection{Observations and Comparisons}
The models displayed clear differences in their handling of harmful content. GPT offered a strong balance, combining effective detection with a relatively low error rate. This made it suitable for environments where both safety and user experience matter. Claude’s high precision and low false positive rate reflect a more cautious approach, making it better for contexts where false accusations must be minimized. Gemini, with the highest recall, is more aggressive, potentially fitting platforms that prioritize safety even at the cost of occasional over-flagging.

Each model reflects a distinct moderation philosophy. GPT is consistent and balanced, Claude is precise and conservative, and Gemini is proactive and broad-reaching. The choice of model should depend on the platform’s goals and tolerance for errors in either direction.

\subsection{Visual Comparison}
\begin{figure}[h]
\centering
\includegraphics[width=\linewidth]{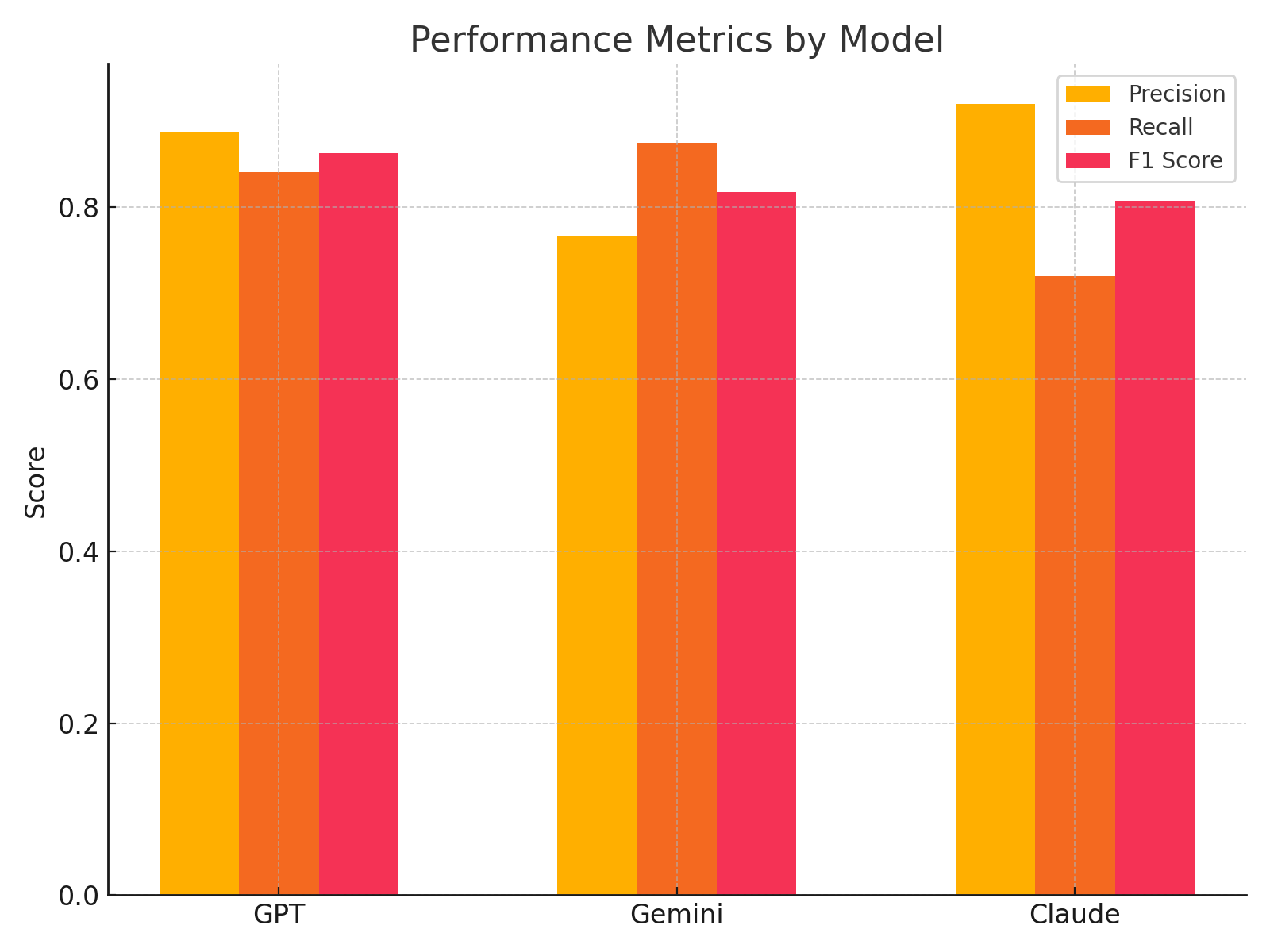}
\caption{Comparison of precision, recall, and F1 score across evaluated models.}
\label{fig:metrics}
\end{figure}

Figure \ref{fig:metrics} compares the models using precision, recall, and F1 score. These metrics highlight how each model prioritizes different aspects of content moderation.

The visual comparison supports earlier findings. GPT maintained a steady balance between precision and recall. Claude achieved top precision but lagged in recall. Gemini led in recall, though with lower precision. These patterns underline that no single model is best across all metrics. Instead, the optimal choice depends on what kind of moderation a platform values most. A platform that can tolerate more false positives may prefer Gemini, while one focused on minimizing moderation mistakes may lean toward Claude. In higher-risk environments, a combination of models could help match the tone, topic, and user profile of each comment more effectively.

\section{Discussion}

\subsection{False Positives and Surface-Level Flagging}\label{sec:falsepositive}
To better understand model over-flagging, we analyzed cases where AI models marked comments as harmful while human reviewers did not. These false positives were grouped by model agreement patterns to highlight shared tendencies and specific weaknesses. While most flagged content reflected strong language or emotional tone, it often lacked any abusive intent. Below are four common patterns observed.

\subsubsection{All Three Models vs. Human: Shared Over-flagging Patterns}
In a small number of cases, only 12 out of 5\,080, all three models (GPT, Claude, and Gemini) identified comments as harmful, while human reviewers did not. These examples typically involved informal phrasing, sarcasm, or emotionally expressive language. For instance, the comment ``Y’all are so toxic'' was likely flagged due to tone, despite not targeting a specific individual. Another example, ``Omg...321lbs! That’s so sad,'' lacked context and may have been flagged merely for mentioning weight. These instances illustrate how all three models tend to rely on surface-level cues, such as slang or emotional keywords, without fully assessing intent or context.

\subsubsection{GPT: Over-Flagging Emotional/Mental-Health Language}
GPT’s false positives often involved emotionally charged or mental health-related phrasing that human reviewers did not consider harmful. These included expressions of concern, frustration, or informal critique that lacked abusive intent. GPT frequently flagged comments referencing emotional breakdowns, support for recovery, or dramatic language about personal change. While this may reflect a cautious stance toward sensitive topics, it also suggests an overreliance on trigger phrases without accounting for context or tone.

\subsubsection{Gemini: Over-Flagging Concern and Instability}
Gemini frequently flagged comments that referenced mental health, personal change, or emotional concern, even when human reviewers found no clear harm. Many of these comments offered support, expressed worry, or questioned the authenticity of a user’s behavior in dramatic or sarcastic terms. Phrases like ``please take care of yourself'' or ``you can’t keep scaring everyone like this'' were often interpreted by Gemini as harmful. This suggests a tendency to over-flag emotionally sensitive or speculative language, particularly when it touches on perceived instability or trauma.

\subsubsection{Claude: Over-Flagging Sarcasm and Hyperbole}
Claude’s false positives often involved sarcastic remarks, informal roasts, or exaggerated critiques that lacked targeted hostility. Many flagged comments included casual profanity, meme references, or expressive language, for example, jokes about appearance, pop culture, or emotional overreactions. In several cases, Claude appeared to react to tone or strong phrasing rather than the presence of actual harm. This suggests that Claude may place greater weight on civility and politeness, leading it to flag socially edgy but benign content more frequently than human reviewers.

These patterns align with Table~\ref{tab:metrics}: Claude’s high precision shows caution, Gemini’s lower precision reflects aggressiveness, and GPT sits between.

\subsection{False Negatives: Sarcasm, Subtext, and Missed Harm}

\subsubsection{All Models vs. Human: Missed Sarcasm, Mockery, and Coded Harassment}
Several harmful comments were missed by all three models, highlighting common blind spots around sarcasm, ridicule, and indirect hostility. Many of these remarks used emojis, mock praise, or exaggerated phrasing to insult, humiliate, or incite collective disdain. Examples included references to ``clown emojis,'' indirect threats like ``he feeds off his haters, just ignore him to make him fall off,'' or comparisons to breakdowns and mental illness framed as jokes.

\subsubsection{GPT: Missed Hostility Behind Humor}
GPT missed a substantial number of harmful comments flagged by human reviewers due to sarcasm, coded ridicule, or indirect hostility. Phrases like ``every time I see you in the hospital I smile'' conveyed sustained mockery and dismissiveness toward someone’s well-being. These misses suggest that GPT may rely too heavily on surface-level tone and literal phrasing.

\subsubsection{Gemini: Missed Harm in Public Shaming}
Gemini failed to flag a range of comments often involving mockery or aggression wrapped in sarcasm or cultural shorthand. Some escalated into public shaming, such as calling for mass unsubscribing or questioning the person’s sanity. Gemini's tendency to underreact suggests it may struggle with patterns of collective ridicule.

\subsubsection{Claude: Missed Sarcasm and Faux Concern}
Claude frequently overlooked comments like ``Send this man back to Arkham,'' which ridicule the creator through implication. It also failed to detect hostile rants accusing the creator of deception or instability, showing a reluctance to flag comments unless explicitly hostile.

\subsection{Risks of Missed Moderation in Sensitive Cases}
False negatives are not equal in impact. While a single missed insult may seem minor, its harm can compound over time, especially in emotionally vulnerable environments. Several missed comments in this study targeted known vulnerabilities, mocked prior hospitalization, or questioned a creator’s mental health. These remarks, though subtle or sarcastic, reinforce harmful narratives and may discourage affected users from seeking help or participating in the platform.

More troubling were coordinated suggestions to ``unfollow so he falls off'' or ``ignore him until he cracks again.'' These comments reflect collective hostility rather than isolated aggression. When such messages go unflagged, they enable mob harassment campaigns that can lead to reputational harm, social withdrawal, or emotional distress for creators. In several cases, these remarks appeared alongside strings of other taunts, suggesting that their impact is not just in their wording but in their cumulative pressure.

This highlights a key limitation of single-comment moderation: by treating each comment in isolation, models miss larger patterns of piling-on or manipulation. Moderation systems must evolve to track conversation history, detect repeated targeting, and recognize coded language that signals coordinated behavior. Especially in cases involving mental health, even a few missed comments can shift the tone of a thread and undermine user safety. Future systems should include tools for temporal tracking, conversational memory, and risk-tiered escalation to reduce the long-term impact of missed moderation.

\subsection{Humor, Satire, and Ambiguity in Model Moderation}

Another common source of disagreement between human reviewers and LLMs involved ambiguous comments that used humor, satire, or double meaning. These false positives often lacked direct hostility but contained emotionally charged phrasing, exaggerated tone, or social critique. Several examples reveal how models misinterpret stylistic or cultural cues.

One such comment was ``Can't sing for nothing!!!!'', flagged by GPT despite being a common form of exaggerated critique in entertainment contexts. Similarly, ``2:33 too much jiggling'' was flagged by Gemini, likely due to its focus on physical attributes, though the intent may have been commentary rather than harassment.

Comments referencing public figures or using layered sarcasm were also commonly misclassified. For instance, ``It sounds like UserNameProtected Cobain's remembrance video'' was flagged by GPT, Gemini, and Claude, though human reviewers judged it to be a stylistic comparison rather than an attack. Another flagged comment, ``You just gonna act like nothing happened?...'' shows how models can mistake rhetorical or skeptical questions for aggression, especially when stripped of conversational context.

Claude and Gemini in particular flagged remarks like ``UserNameProtected ain't even real'' that are commonly used in meme culture or satire. These comments, while edgy, were not interpreted as harmful by human reviewers. However, the models appeared sensitive to tone and phrasing that hinted at ridicule or disbelief.

These cases underscore the limits of current models in interpreting intent, especially when humor blurs the line between critique and cruelty. Without access to conversational history or platform-specific norms, LLMs often make conservative judgments, flagging emotionally charged or socially playful content as harmful.

For moderation systems, this presents a difficult trade-off. Over-flagging benign humor risks alienating users and undermining trust in automated systems, while under-flagging allows veiled insults or passive aggression to persist. Future moderation pipelines may benefit from humor-aware classifiers or community-tuned thresholds that distinguish cultural satire from actual abuse.

\section{Future Work}

The open issues are ranked below from highest near-term impact to longer-term research challenges.

\begin{enumerate}

\item Thread-level and multimodal context.  
      Adding preceding turns, video transcripts, or image cues is the fastest way to cut many sarcasm and piling-on errors found in this study.

\item Ensemble and risk-tiered pipelines.  
      A two-stage approach (first a high-recall filter, then a high-precision check) can be deployed with existing APIs. Measuring latency, cost, and user impact belongs in the next round of experiments.

\item Implicit and coded abuse corpora.  
      New datasets that label irony, metaphor, and insider slang will let researchers test prompt tweaks and specialised pre-training for covert hostility. This requires a dedicated annotation effort.

\item User-facing feedback loops.  
      Controlled trials comparing silent removal, warning prompts, and educational pop-ups could clarify which intervention nudges commenters toward civility most effectively.

\item Long-term mental-health outcomes.  
      Ultimately we need evidence that better moderation lowers anxiety, self-harm ideation, or community churn. Partnering with mental-health researchers and analysing de-identified longitudinal data are goals for later phases.

\end{enumerate}

\section{Conclusion}
This paper benchmarks OpenAI GPT-4.1, Google Gemini 1.5 Pro, and Anthropic Claude 3 Opus on 5\,080 YouTube comments drawn from high-abuse threads in gaming, lifestyle, food vlog, and music channels. Each model was evaluated with the same prompt under deterministic settings. Gemini identified the largest share of harmful content, achieving recall of 0.875, yet its precision dropped to 0.767 because of frequent false positives. Claude reached precision of 0.920 and the lowest false-positive rate of 0.022, although its recall fell to 0.720. GPT-4.1 delivered the best overall balance with an F1 score of 0.863, precision of 0.887, and recall of 0.841. Qualitative analysis showed that sarcasm, coded insults, and mixed-language slang remain persistent blind spots for all three models. These findings make clear that no single system meets every moderation requirement. Practical pipelines should combine complementary models, incorporate conversational context, and fine-tune for under-represented languages and subtle forms of abuse.

For deployment, platforms can map each model’s strengths to specific risk tiers. A site that values maximum coverage can run Gemini as the first-pass filter and then route its flags to Claude or a human queue to reduce false alarms. Low-latency chat services may prefer using GPT alone for a balanced trade-off between recall and precision. Logging each disagreement and feeding it into periodic retraining supports continuous improvement, auditability, and fairness, contributing to AI-safety goals that limit user exposure to harmful content without over-silencing benign speech.

The fully de-identified dataset, prompt text, and model outputs are openly released to support reproducible research and to foster further progress in automated content moderation.

\section{Limitations}
This study has several limitations:

\begin{enumerate}
  \item Zero-shot prompts: All models were evaluated using identical zero-shot prompts without any model-specific fine-tuning or prompt optimization. This setup maximizes comparability but may understate each model’s peak performance.
  \item Comment-level context: Each comment was assessed in isolation, with no access to preceding messages or thread history. As a result, context-dependent abuse, sarcasm, or coordinated behavior across comments may go undetected.
  \item Cultural bias in labeling: Human annotations came from two reviewers sharing similar linguistic and cultural backgrounds, which could bias judgments of sarcasm, slang, or ambiguous phrasing.
  \item Limited language scope: Although a few comments were in Arabic or Indonesian, the dataset is overwhelmingly English. Therefore, our findings should not be generalized to truly multilingual moderation performance.
  \item No BERT-based baseline: Traditional transformer classifiers like BERT and RoBERTa have already been widely studied in toxic language detection tasks \cite{b17, b18}. This work focuses instead on production-grade LLMs that are currently deployed or considered for content moderation at scale. Including legacy baselines would provide limited additional insight relative to the study’s primary goal of evaluating state-of-the-art zero-shot LLM performance.

\end{enumerate}

\section*{Ethical Considerations}
This study uses only publicly available YouTube comments. All user identifiers and any personally identifying details were removed or replaced with neutral placeholders before analysis, so no individual can be re-identified. Because the data are de-identified and pose minimal risk, no formal Institutional Review Board (IRB) review was required. 

To address the cultural-bias risk noted in the Limitations section, future releases will include reviews from annotators outside the authors’ demographic group and will make the labeling guide public so external researchers can audit or challenge specific decisions.

\section*{Data Availability}
The full, de-identified comment corpus, together with the human labels and model predictions, is openly available at \url{https://github.com/Ammce/papers/tree/main/llm-cyberbullying-moderation%20}. To minimize re-identification risk, user names, links, and YouTube comment IDs have been removed.  The list of source‐video IDs can be shared with editors or qualified researchers under a non-disclosure agreement.  Exact model prompts and API parameters used in this study are provided in the same repository under \texttt{prompts/}.


\begin{thebibliography}{00}

\bibitem{b1}
B. Dean, ``Social media usage \& growth statistics,'' \emph{Backlinko}, Feb. 21, 2024. [Online]. Available: https://backlinko.com/social-media-users.

\bibitem{b2}
A. B. Barragán Martín \emph{et al.}, ``Study of cyberbullying among adolescents in recent years: A bibliometric analysis,'' \emph{Int. J. Environ. Res. Public Health}, vol. 18, no. 6, p. 3016, Mar. 2021. [Online]. Available: https://doi.org/10.3390/ijerph18063016.

\bibitem{b3}
S. Hinduja and J. W. Patchin, ``Bullying, cyberbullying, and suicide,'' \emph{Arch. Suicide Res.}, vol. 14, no. 3, pp. 206–221, 2010. doi:10.1080/13811118.2010.494133.

\bibitem{b4}
C. P. Barlett, ``Anonymously hurting others online: The effect of anonymity on cyberbullying frequency,'' \emph{Psychol. Pop. Media Cult.}, vol. 4, no. 2, pp. 70–79, 2015. doi:10.1037/a0034335.

\bibitem{b5}
L. Huang \emph{et al.}, ``The severity of cyberbullying affects bystander intervention among college students: The roles of feelings of responsibility and empathy,'' \emph{Psychol. Res. Behav. Manag.}, vol. 16, pp. 893–903, Mar. 2023. doi:10.2147/PRBM.S397770.

\bibitem{b6}
A. Vigderman, ``Cyberbullying: Twenty crucial statistics for 2024,'' \emph{Security.org}, Oct. 9, 2024. [Online]. Available: https://www.security.org/resources/cyberbullying-facts-statistics/.

\bibitem{b7}
W. Craig \emph{et al.}, ``Social media use and cyber-bullying: A cross-national analysis of young people in 42 countries,'' \emph{J. Adolesc. Health}, vol. 66, no. 6, pp. S100–S108, Jun. 2020. doi:10.1016/j.jadohealth.2020.03.006.

\bibitem{b8}
M. H. Ribeiro, J. Cheng, and R. West, ``Automated content moderation increases adherence to community guidelines,'' in \emph{Proc. ACM Web Conf. (WWW)}, 2023, pp. 2666–2676. doi:10.1145/3543507.3583275.

\bibitem{b9}
S. Wang and K. J. Kim, ``Content moderation on social media: Does it matter who and why moderates hate speech?'' \emph{Cyberpsychol. Behav. Soc. Netw.}, vol. 26, no. 7, pp. 527–534, Jul. 2023. doi:10.1089/cyber.2022.0158.

\bibitem{b10}
T. Gillespie, ``Content moderation, AI, and the question of scale,'' \emph{Big Data Soc.}, vol. 7, no. 2, pp. 1–5, Jul. 2020. doi:10.1177/2053951720943234.

\bibitem{b11}
H. Lopez and S. Kübler, ``Context in abusive language detection: On the interdependence of context and annotation of user comments,'' \emph{Discourse, Context Media}, vol. 63, Art. no. 100848, Feb. 2025. doi:10.1016/j.dcm.2024.100848.

\bibitem{b12}
M. van Geel, P. Vedder, and J. Tanilon, ``Relationship between peer victimization, cyberbullying, and suicide in children and adolescents: A meta-analysis,'' \emph{JAMA Pediatr.}, vol. 168, no. 5, pp. 435–442, May 2014. doi:10.1001/jamapediatrics.2013.4143.

\bibitem{b13}
Z. Waseem and D. Hovy, ``Hateful symbols or hateful people? Predictive features for hate speech detection on Twitter,'' in \emph{Proc. NAACL Student Res. Workshop}, San Diego, CA, USA, Jun. 2016, pp. 88–93. doi:10.18653/v1/N16-2013.

\bibitem{b14}
A.-M. Founta \emph{et al.}, ``Large scale crowdsourcing and characterization of Twitter abusive behavior,'' in \emph{Proc. Int. Conf. Web Social Media}, Atlanta, GA, USA, Mar. 2018, pp. 491–500. doi:10.1609/icwsm.v12i1.14991.

\bibitem{b15}
M. Zampieri \emph{et al.}, ``Predicting the type and target of offensive posts in social media,'' in \emph{Proc. NAACL}, Minneapolis, MN, USA, Jun. 2019, pp. 1415–1420. doi:10.18653/v1/N19-1144.

\bibitem{b16}
P. Röttger, B. Vidgen, D. Nguyen, Z. Waseem, H. Margetts, and J. Pierrehumbert, ``HateCheck: Functional tests for hate speech detection models,'' in \emph{Proc. 59th Annu. Meet. Assoc. Comput. Linguistics \& 11th Int. Joint Conf. NLP (Long Papers)}, Online, Aug. 2021, pp. 41–58. doi:10.18653/v1/2021.acl-long.4.

\bibitem{b17}
J. Devlin, M. Chang, K. Lee, and K. Toutanova, ``BERT: Pre-training of deep bidirectional transformers for language understanding,'' in \emph{Proc. NAACL-HLT}, Minneapolis, MN, USA, Jun. 2019, pp. 4171–4186. doi:10.18653/v1/N19-1423.

\bibitem{b18}
Y. Liu \emph{et al.}, ``RoBERTa: A robustly optimized BERT pretraining approach,'' arXiv preprint arXiv:1907.11692, Jul. 2019. [Online]. Available: https://arxiv.org/abs/1907.11692.

\bibitem{b19}
B. Mathew, P. Saha, S. M. Yimam, C. Biemann, P. Goyal, and A. Mukherjee, ``HateXplain: A benchmark dataset for explainable hate speech detection,'' in \emph{Proc. AAAI Conf. Artif. Intell.}, vol. 35, no. 17, May 2021, pp. 14867–14875. doi:10.1609/aaai.v35i17.17745.

\bibitem{b20}
B. Vidgen, T. Thrush, Z. Waseem, and D. Kiela, ``Learning from the worst: Dynamically generated datasets to improve online hate detection,'' in \emph{Proc. 59th Annu. Meet. Assoc. Comput. Linguistics \& 11th Int. Joint Conf. NLP (Long Papers)}, Aug. 2021, pp. 1667–1682. doi:10.18653/v1/2021.acl-long.132.

\bibitem{b21}
S. Gehman, S. Gururangan, M. Sap, Y. Choi, and N. A. Smith, ``RealToxicityPrompts: Evaluating neural toxic degeneration in language models,'' in \emph{Findings Assoc. Comput. Linguistics: EMNLP 2020}, Nov. 2020, pp. 3356–3369. doi:10.18653/v1/2020.findings-emnlp.301.

\bibitem{b22}
A. Arora, ``Sarcasm detection in social media: A review,'' in \emph{Proc. Int. Conf. Innov. Comput. Commun. (ICICC)}, Dec. 2020, pp. 1–4. doi:10.2139/ssrn.3749018.

\bibitem{b23}
M. S. Jahan and M. Oussalah, ``A systematic review of hate speech automatic detection using natural language processing,'' \emph{Neurocomputing}, vol. 546, Art. no. 126232, Aug. 2023. doi:10.1016/j.neucom.2023.126232.

\bibitem{b24}
J. M. Pérez \emph{et al.}, ``Assessing the impact of contextual information in hate speech detection,'' \emph{IEEE Access}, vol. 11, pp. 30575–30590, 2023. doi:10.1109/ACCESS.2023.3258973.

\bibitem{b25}
H. Mubarak, K. Darwish, and W. Magdy, ``Abusive language detection on Arabic social media,'' in \emph{Proc. 1st Workshop on Abusive Language Online}, Vancouver, Canada, 2017, pp. 52–56. doi:10.18653/v1/W17-3008.

\bibitem{b26}
T. Mandl \emph{et al.}, ``Overview of the HASOC track at FIRE 2019: Hate speech and offensive content identification in Indo-European languages,'' in \emph{Proc. FIRE}, Kolkata, India, 2019, pp. 14–17. doi:10.1145/3368567.3368584.

\bibitem{b27}
T. Gröndahl, L. Pajola, M. Juuti, M. Conti, and N. Asokan, ``\textquotesingle All you need is Love\textquotesingle: Evading hate speech detection,'' in \emph{Proc. 11th ACM Workshop Artif. Intell. Security}, Toronto, Canada, 2018, pp. 2–12. doi:10.1145/3270101.3270103.

\bibitem{b28}
N. Murikinati, A. Anastasopoulos, and G. Neubig, ``Transliteration for cross‐lingual morphological inflection,'' in \emph{Proc. 17th SIGMORPHON Workshop Computational Research Phonetics, Phonology, and Morphology}, Online, Jul. 2020, pp. 189–197. doi:10.18653/v1/2020.sigmorphon-1.22.

\bibitem{b29}
J. Khanuja, A. Dandapat, A. Srinivasan, S. Sitaram, and M. Choudhury, ``GLUECoS: An evaluation benchmark for code‐switched NLP,'' in \emph{Proc. ACL‐IJCNLP}, Bangkok, Thailand, 2021, pp. 3575–3585. doi:10.18653/v1/2020.acl-main.329.

\bibitem{b30}
J. Ranasinghe and M. Zampieri, ``Multilingual offensive language identification with cross‐lingual embeddings,'' in \emph{Proc. EMNLP}, Online, 2020, pp. 5838–5844. doi:10.18653/v1/2020.emnlp-main.470.

\bibitem{b31}
Ç. Çöltekin, ``A corpus of Turkish offensive language on social media,'' in \emph{Proc. LREC}, Marseille, France, 2022, pp. 4878–4885. [Online]. Available: https://aclanthology.org/2020.lrec-1.758/.

\bibitem{b32}
E. Pamungkas and V. Patti, ``Cross‐domain and cross‐lingual abusive language detection: A hybrid approach with deep learning and a multilingual lexicon,'' in \emph{Proc. ACL}, Florence, Italy, 2019, pp. 363–370. doi:10.18653/v1/P19-1051.

\bibitem{b33}
Y. Liu and M. Zhang, ``LLM‐Mod: Can Large Language Models Assist Content Moderation?'' in \emph{Proc. ACM Conf. Fairness, Accountability, and Transparency (FAccT)}, Rio de Janeiro, Brazil, 2024, pp. 1–12. doi:10.1145/3613905.3650828.

\bibitem{b34}
F. M. Plaza‐Del‐Arco, D. Nozza, and D. Hovy, ``Respectful or toxic? Using zero‐shot learning with language models to detect hate speech,'' in \emph{Proc. 7th Workshop Online Abuse and Harms (WOAH)}, Singapore, Jan. 2023, pp. 46–52. doi:10.18653/v1/2023.woah-1.6.

\bibitem{b35}
J. Pavlopoulos \emph{et al.}, ``Toxicity detection: Does context really matter?'' in \emph{Proc. ACL}, Online, 2020, pp. 4296–4305. doi:10.18653/v1/2020.acl-main.396.

\bibitem{b36}
A. Baheti, M. Sap, and Y. Tsvetkov, ``Just say no: Analyzing the stance of neural dialogue generation in offensive contexts,'' in \emph{Proc. EMNLP}, Online, 2021, pp. 4846–4859. doi:10.18653/v1/2021.emnlp-main.397.

\bibitem{b37}
M. Sap \emph{et al.}, ``Social bias frames: Reasoning about social and power implications of language,'' in \emph{Proc. ACL}, Online, 2020, pp. 5477–5490. doi:10.18653/v1/2020.acl-main.486.

\bibitem{b38}
I. Solaiman and C. Dennison, ``Process for adapting language models to society (PALMS),'' Tech. Rep., OpenAI, 2021. [Online]. Available: https://arxiv.org/abs/2106.10328.

\bibitem{b39}
T. Bolukbasi, K.-W. Chang, J. Zou, V. Saligrama, and A. Kalai, ``Man is to computer programmer as woman is to homemaker? Debiasing word embeddings,'' in \emph{Proc. NeurIPS}, Barcelona, Spain, 2016, pp. 4356–4364. doi:10.48550/arXiv.1607.06520.

\bibitem{b40}
H. Welbl, A. Stiennon, and Y. Bai, ``Challenges in detoxifying language models,'' Tech. Rep., DeepMind, 2021. [Online]. Available: https://arxiv.org/abs/2109.07445.

\bibitem{b41}
R. Hartvigsen, H. Palangi, and X. He, ``Toxigen: Controllable generation of implicit and adversarial toxic text,'' in \emph{Proc. ACL}, Dublin, Ireland, 2022, pp. 524–535. doi:10.18653/v1/2022.acl-long.39.

\bibitem{b42}
E. Bender \emph{et al.}, ``On the dangers of stochastic parrots,'' in \emph{Proc. FAccT}, Online, 2021, pp. 610–623. doi:10.1145/3442188.344592.

\end{thebibliography}
\end{document}